# YH-MINER: Multimodal Intelligent System for Natural Ecological Reef Metric Extraction

**Mingzhuang Wang [a, c†], Yvyang Li [b†], Xiyang Zhang [a], Fei Tan [a], Qi Shi [a], Guotao Zhang [a, c], Siqi Chen [a, c], Yufei Liu [d], Lei Lei [e], Ming Zhou [e], Qiang Lin [a], Hongqiang Yang [a*]**

[a] South China Sea Institute of Oceanology, Chinese Academy of Sciences, Guangzhou 510301, China

[b] National Astronomical Observatories of the Chinese Academy of Sciences, Beijing 100012, China

[c] University of Chinese Academy of Sciences, Beijing 100049, China

[d] International Department of Affiliated High School of South China Normal University, Guangzhou 510630, China

[e] Alibaba Cloud, Hangzhou 310051, China

**† Contributed equally to this work**

*** Corresponding author:** Hongqiang Yang

**Email address:** hqyang@scsio.ac.cn

**Abstract**

Coral reefs, crucial for sustaining marine biodiversity and ecological processes (e.g., nutrient cycling, habitat provision), face escalating threats, underscoring the need for efficient monitoring. Coral reef ecological monitoring faces dual challenges of low efficiency in manual analysis and insufficient segmentation accuracy in complex underwater scenarios. This study develops the YH-MINER system, establishing an intelligent framework centered on the Multimodal Large Model (MLLM) for "object detection-semantic segmentation-prior input". The system uses the object detection module (mAP@0.5=0.78) to generate spatial prior boxes for coral instances, driving the segment module to complete pixel-level segmentation in low-light and densely occluded scenarios. The segmentation masks and finetuned classification instructions are fed into the Qwen2-VL-based multimodal model as prior inputs, achieving a genus-level classification accuracy of 88% and simultaneously extracting core ecological metrics. Meanwhile, the system retains the scalability of the multimodal model through standardized interfaces, laying a foundation for future integration into multimodal agent-based underwater robots and supporting the full-process automation of "image acquisition-prior generation-real-time analysis".

## 1. Introduction

As one of the most biologically diverse marine ecosystems on Earth, coral reefs are renowned as the "tropical rainforests of the ocean" due to their complex three-dimensional structures and efficient material-energy cycling mechanisms (Rajasuriya, Öhman et al., 1997; Ferrario et al., 2014; Steneck et al., 2018; Woodhead et al., 2019; Shumway, Foster, and Fidelman 2025). Despite covering only 0.2% of the ocean's area, these ecosystems sustain 25% of global marine biodiversity, directly or indirectly providing multiple services for over 1 billion people, including food security, coastal protection, and tourism economies (Sing Wong, Vrontos et al., 2022). However, under the dual pressures of global climate change and human activities, coral reefs are experiencing an unprecedented degradation crisis. Large-scale coral bleaching events occur repeatedly at the global spatial scale. Coral mortality caused by bleaching triggers abrupt changes in reef structure, biodiversity, productivity, and functionality. Such ecological catastrophes not only lead to the functional loss of Scleractinia species but also threaten reef-associated organisms with habitat fragmentation (Graham et al., 2015; Hughes et al., 2017; Hughes, Kerry, et al., 2018; Hughes, Anderson, et al. ,2018; Eddy et al., 2021). Monitoring data from the South China Sea show that coral reefs in this region are generally degrading, with significant declines in live coral cover and reductions in the number of reef-building coral genera, severely threatening regional marine ecological security (Yu 2012; Chen et al., 2009; Wong et al., 2018).

Coral quadrat surveys provide critical ecological information, including coral coverage, species richness, species composition, and community structure (Perera-Valderrama et al., 2016; Sato et al., 2018; Lechene et al., 2019). As standardized tools for ecological monitoring, quadrats quantify coral data within unit areas, reflecting not only the structural status of coral reef ecosystems (e.g., live coral cover percentage) but also revealing their dynamic change patterns (Sato et al., 2018). These data are scientifically valuable for assessing reef health, predicting community succession trends, and quantifying the impacts of environmental disturbances. Accurately quantifying core indicators of coral reef ecological health is a key scientific foundation for formulating effective conservation strategies. Live coral cover, as a core parameter reflecting ecosystem structural integrity, enables in-depth understanding of climate change impacts and reef development trends through monitoring (Jiang, Qu et al., 2023). Meanwhile, species richness and diversity indices of coral taxonomic composition reveal ecosystem resilience potential (Carturan, Parrott et al., 2022). Together, live coral cover and species composition constitute the core parameters of coral reef ecological monitoring, revealing ecosystem health status, adaptive potential, and conservation needs from the perspectives of "quantity" and "quality," respectively.

Traditional methods for extracting coral quadrat data (e.g., coral cover calculation, taxonomic identification, species diversity assessment) highly depend on expert experience and manual visual interpretation, suffering from limitations such as time-consuming processes, high costs, and subjective errors. With the rapid development of artificial intelligence, deep learning-based automated analysis methods have gradually become key tools to overcome these bottlenecks. Deep learning has emerged as a pivotal tool for automating coral reef monitoring, with convolutional neural networks (CNNs) and transformer-based architectures increasingly employed to address challenges in image analysis (Gómez-Ríos et al., 2019; Wang et al., 2020). However, traditional models face persistent bottlenecks in handling low-resolution imagery, data imbalance, and cross-domain variability—critical issues in underwater coral photography (Nguyễn et al., 2021; Rivas et al., 2023). Due to data or model issues, recent advancements in transfer learning, such as EfficientNet (Tan and Le 2019) and Swin Transformer (Liu et al., 2021), have significantly improved general feature representation through large-scale pre-training on datasets like ImageNet (Deng et al., 2009) and CoCo (Lin et al., 2014), enhancing model performance and robustness across diverse environmental conditions. Building on this, MLLMs, pre-trained on essentially all existing high-quality image datasets, and even a large amount of cleaned internet data, demonstrate superior capabilities in integrating visual semantics—including implicit familiarity with coral-like structures from indirect data exposure (For example: What are the characteristics and distribution of the genus Acropora? Answer: Acropora is characterized by its branching or tabular growth forms, rapid growth rate, and high ecological importance as primary reef builders. It primarily distributes in tropical and subtropical shallow marine regions of the Indo-Pacific, Red Sea, and Caribbean, though some species show tolerance to varying water temperatures and turbidity). Nevertheless, their performance remains constrained by "hallucination" risks arising from low-quality labels or resolution mismatches in pre-training data, particularly for fine-grained coral taxa (For example: Which corals are included in this image, what is the dominant genus, and what is the diversity index? Answer: This question contains errors. Without specific image input or field survey data, it is impossible to accurately identify coral species, determine dominant genera, or calculate diversity indices. Such analyses require validated image datasets or standardized ecological sampling, and speculative answers may introduce erroneous conclusions).

To explore the possibility of precise coral identification and auxiliary accurate coverage estimation, we aim to develop a framework based on broad-domain pre-training and fine-tuning with high-quality coral data. We have compiled a cross-regional high-quality dataset containing 43 major coral genera. Image segmentation results are used as spatial priors to assist model recognition. However, semantic segmentation performance is constrained by sample specificity, underwater environmental noise, and quadrat occlusion in complex scenes. To address these challenges, object detection is introduced to enhance few-shot learning capabilities, particularly for sparse or occluded coral instances.

Thus, To address these challenges, we propose a novel framework within the YaoHua (YH) series of coral large models, integrating object detection, image segmentation, and MLLM-based coral recognition to achieve more accurate data extract. Specifically, the object detection module first localizes coral instances in underwater images, generating bounding boxes to overcome the limitations of semantic segmentation in low-quality or cluttered scenarios. The segmentation module then produces pixel-level masks using these bounding boxes as prompts, providing fine-grained spatial information. These masks, combined with visual features from coral images, are fed into a MLLM for genus-level classification, enabling hierarchical extraction of ecological metrics (e.g., coral cover, biodiversity indices). The modular design of the framework, with standardized interfaces of the MLLM, lays a foundation for future integration with multimodal agent-based underwater robotic systems, facilitating real-time on-reef monitoring and dynamic ecological assessment (Liu et al., 2014).

This paper is structured to systematically present the development and evaluation of the YH-MINER

system. Section 2 presents a comprehensive overview of the Coral Quadrat Intelligent Analysis System, with a focus on its three core modules: (1) the object detection-segmentation module employing a cascaded framework for precise coral instance localization, (2) the classification module integrating a multimodal coral identification model, and (3) the data extraction module enabling hierarchical ecological metric calculation. Following this, Section 3 details the construction of our multi-source coral dataset featuring 114,042 high-resolution images across 43 genera for different tasks. Section 4 provides a rigorous analysis of the training methodology, emphasizing hyperparameter optimization techniques and ablation studies that informed model development. The experimental validation in Section 5 comprehensively evaluates system performance through both quantitative metrics and qualitative analyses of genus-specific confusion patterns. Finally, Section 6 contextualizes our contributions within the broader field of coral reef monitoring and discussed technical limitations related to morphological convergence challenges.

## 2.The YH-MINER System

### 2.1 Object Detection and Segmentation Module

This module integrates an object detection model and an image segmentation model through a cascaded framework to achieve precise localization and semantic segmentation of coral instances. The YOLO11s based (Redmon et al., 2016) detection component employs a trained multi-scale feature fusion mechanism optimized for underwater coral imagery, enabling real-time object detection that outputs spatially accurate bounding boxes. All model parameters participate in training, but due to the adoption of a small learning rate and warm-start strategy, the actual update amount adapts to data characteristics, preferentially optimizing parameters in the detection head and Neck components. Based on these detection results, the system automatically generates spatial prompts (box prompts) to guide the SAM-based (Kirillov et al., 2023) segmentation module in performing pixel-level delineation of coral regions. This cascaded design addresses the challenges of cross-domain adaptation in coral segmentation tasks, demonstrating robust performance across diverse ecological scenarios—from sparse coral colonies (e.g., isolated individuals) to densely packed reef structures (e.g., complex aggregations). The resulting binary segmentation masks provide critical spatial foundations for subsequent ecological metric calculations, including coral cover estimation and population topology analysis. By combining detection-driven spatial priors with segmentation-based fine-grained delineation, the module effectively bridges the gap between macro-scale localization and microstructural precision required for coral reef monitoring.

### 2.2 Cover Recognition and Calculation Module

The Recognition module enhances performance through joint optimization of the visual encoder, projector, and LLM components during training. By integrating text-driven instructions with segmentation-derived spatial priors—where binary masks from the segmentation module act like visual prompt to highlight coral regions—the system improves genus-level identification accuracy while suppressing background interference. This approach directly addresses challenges in distinguishing morphologically similar genera through domain-specific parameter adjustments and spatial constraint guidance.

Calculation Module hierarchically extracts ecological metrics based on segmentation masks and classification results. Relative coral cover is calculated by determining the proportion of coral pixels within the total quadrat area, while detection results correct overlapping regions to eliminate segmentation errors in dense coral aggregations. Coral richness is quantified by statistically analyzing the number of identified coral genera within the quadrat to characterize community composition diversity. Finally, coral abundance (Abundance), Shannon-Wiener index (H'), and Simpson index (D) are computed based on cover and genus distribution to assess community stability and ecological functional potential. The final ecological data are exported in standardized formats (e.g., CSV or JSON), enabling comparative analysis with traditional quadrat survey data and providing quantitative foundations for dynamic coral reef health monitoring.

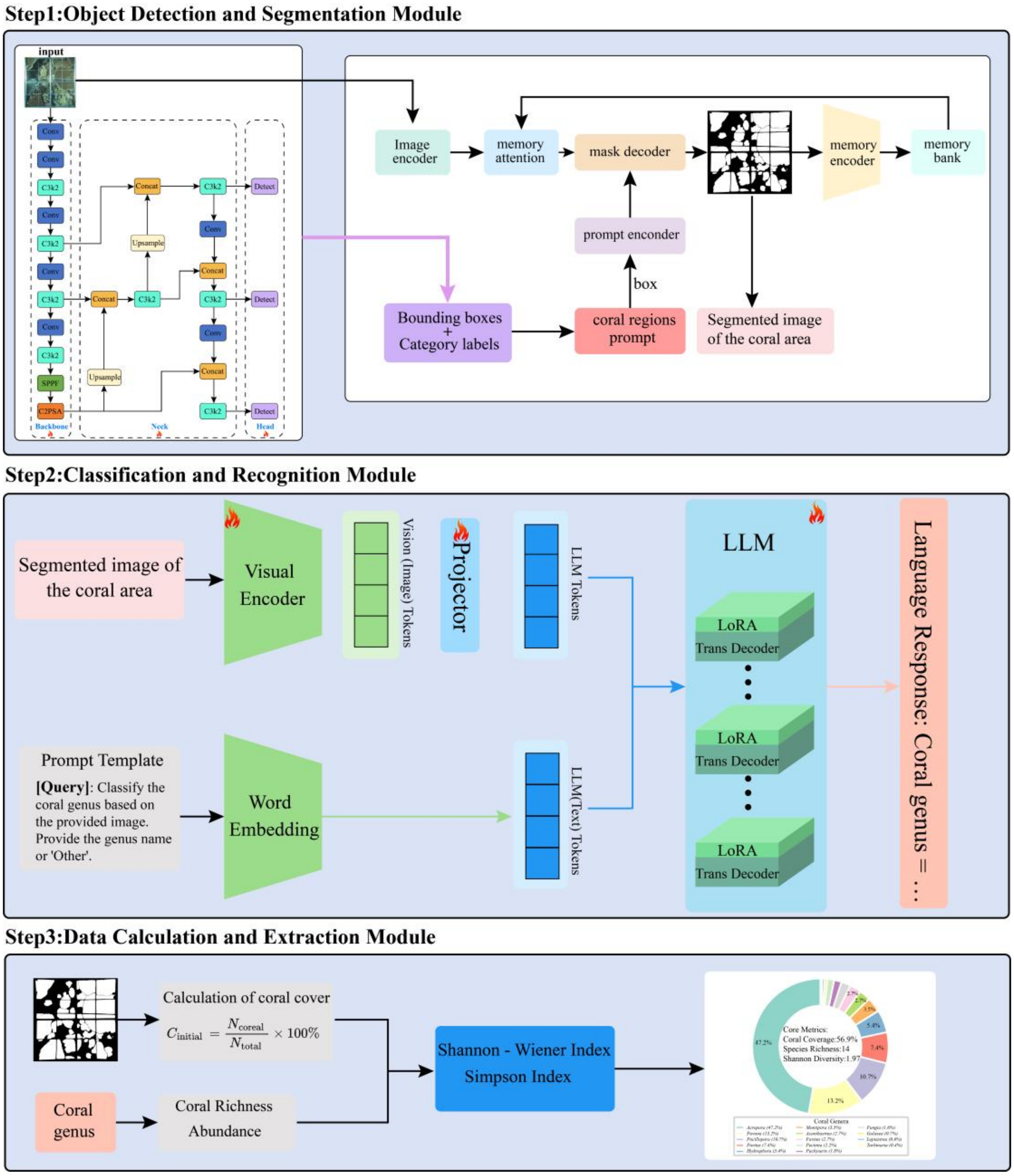

**Figure 1. Smart Coral Quadrat Analysis System (YH-MINER) Structural Diagram**

The first step performs target detection and segmentation of the coral sample, the segmented coral is used as an input to the coral classification and identification module in the second step to identify the coral, and the third step combines the outputs of the first step and the second step to extract the coral sample data. Where $N_{coral}$ the total number of white pixels in the coral area of the coral sample, $N_{total}$ is the total number of pixels in the sample, and initial coverage $C_{initial}$ is the proportion of coral pixels to the total area of the sample. Sparks represent trained.

## 3. Experimental Process

### 3.1 Dataset Construction

This study develops a coral reef intelligent recognition dataset through multi-source data integration, specifically designed to enhance model generalization in ecologically complex scenarios. The dataset is structured into three hierarchically interconnected components: (1) Core Dataset focusing on dominant coral diversity with 9 major genera (*Acropora*, *Porites*, etc.) accounting for 90% of taxonomic representation; (2) Extended Dataset expanded through cross-platform fusion (CoalNet, GBIF, RSMAS/EILAT, Marine Life Photography database, Gómez-Ríos et al., 2019) containing 114,042 images across 43 coral genera at resolutions up to 1024×1024 pixels (see TableA.1); and (3) Object Detection Dataset specialized for localization tasks, covering three ecological scenarios (individual colonies, sparse assemblages, dense aggregations) with spatially calibrated annotations. Through hierarchical screening and cross-source fusion, this framework establishes a standardized benchmark that surpasses traditional datasets (EILAT, EILAT2, RSMAS) in three critical dimensions: taxonomic breadth with 5.4× more genera represented, spatial resolution with 16× higher median image quality preserving microstructural details, and ecological realism with two orders of magnitude larger data volume and expanded spatiotemporal coverage. This enhanced diversity and scale directly address main limitations of conventional benchmarks: insufficient representation of morphologically convergent taxa, resolution constraints in capturing diagnostic skeletal features, and temporal-spatial bias from limited data sources. The resulting dataset enables robust model training for real-world coral reef monitoring, particularly in distinguishing phylogenetically proximate genera through preserved microstructural cues like septal arrangements and coenosteal textures.

The Core Dataset focuses on the dominant portion of coral diversity, comprising the top 2% of coral genera (9 major genera in total) that account for over 90% of the dataset's diversity, ensuring balanced and representative sample distribution. Primarily sourced from public imagery of global coral reef monitoring networks, this dataset covers dominant taxa within the order *Scleractinia*, such as *Acropora* and *Porites*. The distribution of the core dataset is shown in Table 1.

**Table 1. Distribution of coral genera in the Core Dataset**

| Genus | Photo count | Percentage of Total |
|---|---|---|
| *Acropora* | 1772 | 21.72% |
| *Montipora* | 727 | 8.91% |
| *Porites* | 457 | 5.60% |
| *Favia* | 268 | 3.28% |
| *Goniopora* | 194 | 2.38% |
| *Pavona* | 185 | 2.27% |
| *Fungia* | 180 | 2.21% |
| *Pocillopora* | 179 | 2.19% |
| *Leptoseris* | 163 | 2.00% |
| Hybrid | 4035 | 49.44% |

The Extended Dataset significantly enhances model generalization through cross-platform data integration, incorporating four major public resources:

1. CoalNet: Provides continuous monitoring imagery of coral reefs, offering critical spatiotemporal dynamics for ecological process analysis.
2. GBIF: Acquires species distribution records (DOI:10.15468/dl.5zqkph) via its API, covering

taxonomic data for corals across 83 countries globally.

3. RSMAS and EILAT: Specialized datasets for automated coral recognition, containing underwater images under diverse lighting conditions.
4. Public Image Libraries: High-resolution coral images from the Marine Life Photography database (https://www.marinelifephotography.com/corals/corals.htm), supplementing rare taxon samples.

The final Extended Dataset includes over 114,000 images (114,042 total) spanning 43 coral genera and a "Hybrid" category, establishing one of the most comprehensive benchmarks for coral identification (Table A.1). The resolution of coral images in the whole data set is shown in Fig.2.

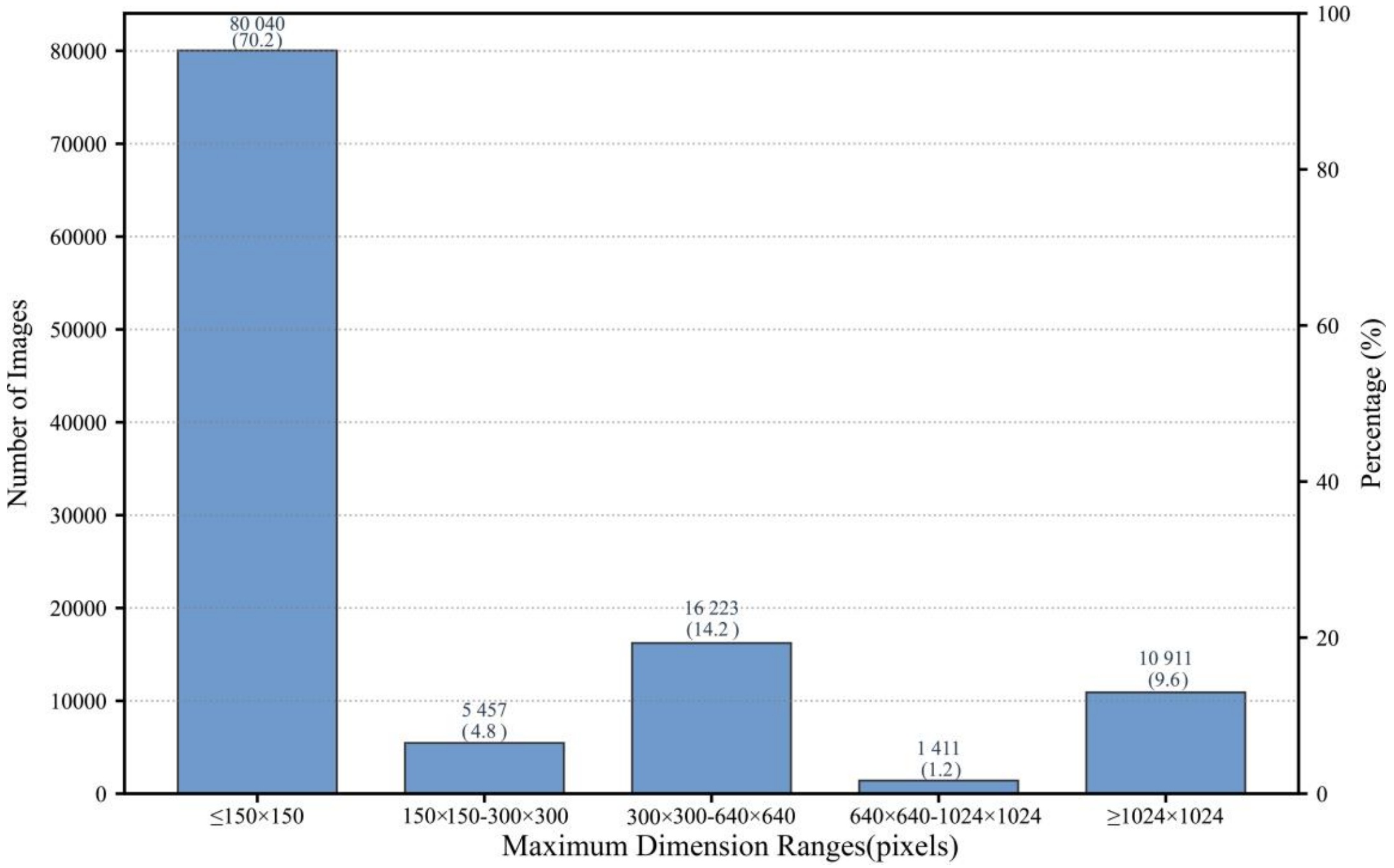

**Figure 2. Resolution histogram of the dataset**

The Object Detection Dataset is specially constructed for object detection model training, which integrates the in-situ coral quadrat images collected by the research group in the South China Sea and the high-quality coral images of the classification data set. It covers three typical ecological scenes and more than 10,000 annotated images:

1. Individual Coral Colonies: Focuses on morphological features of independent coral individuals.
2. Sparse Coral Assemblages: Includes low-density coral-substrate interaction scenarios.
3. Dense Coral Aggregations: Simulates complex spatial structures in natural reef areas.

This multi-scenario coverage strengthens the model's detection capability across different community structures (Fig.3).

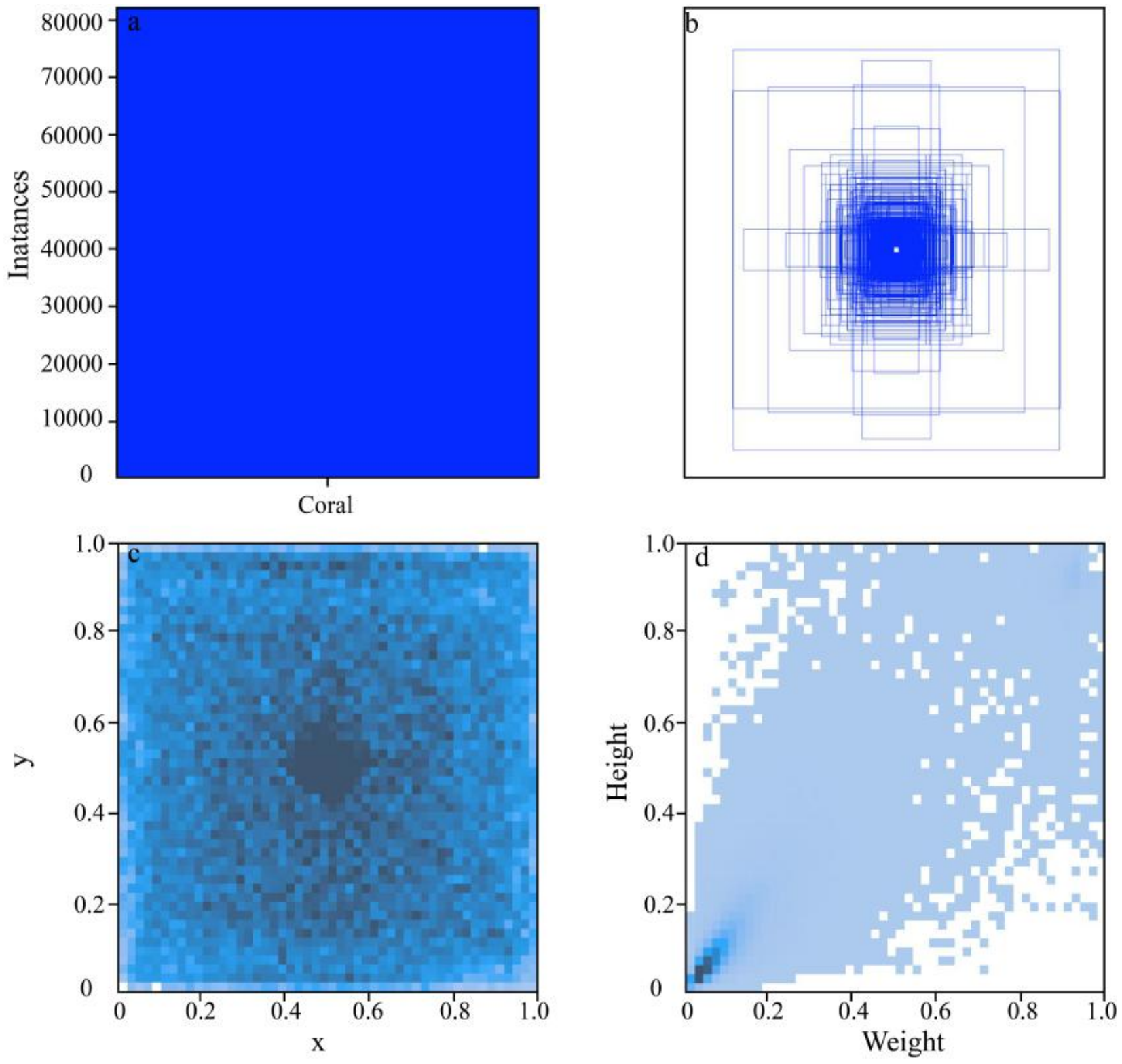

**Figure 3. Object Detection Dataset Analysis**

a: Sample count coral category; b: Spatial distribution of bounding boxes in images; c: Distribution of bounding box center coordinates; d: Aspect ratio and size distribution of detected objects.

### 3.2 Training

#### 3.2.1 Object Detection Part Training

An optimized YOLOv11s architecture was employed for object detection, configured with the AdamW optimizer using an initial learning rate of 0.0005 and weight decay of 0.0005 to mitigate overfitting. To ensure stable training, gradient clipping with a maximum norm of 1 was applied, and a cosine learning rate schedule with a final decay factor of 0.05 was adopted to facilitate smooth convergence. For enhanced small object detection, we increased the input image size to 1024×1024 pixels and extended training epochs, ensuring stable performance on small targets. The model was trained on a single NVIDIA A10 GPU with other parameters adjusted to fit the GPU memory.

#### 3.2.2 Classification Recognition Part Training

We used Qwen2 VL 7B as base models and AdamW as the default optimizer. Gradient clipping with a maximum norm of 1 was used for stable training. A cosine learning rate schedule was adopted to achieve smooth convergence, complemented by a warm-up ratio of 5%. For training the ViT, projector, and LLM components simultaneously, we leveraged LORA technology to reduce memory overhead, using a small number of training epochs with parameters such as lora_rank=64, lora_alpha=32, and lora_dropout=0.01 to control model adaptation. Subsequently, in the full-parameter training phase—with deepspeed-zero2 enabled for memory optimization—the training epochs were increased to 3 using an extended dataset, while moderately lowering the resolution. A control test was conducted during full-parameter training by freezing the ViT part to evaluate its impact on model performance. All experiments were executed on an Ubuntu 20.04 server equipped with 8×Nvidia H20 GPUs.

## 4. Results

### 4.1 Object detection model training results

The object detection model demonstrated robust convergence during training on the coral detection dataset (Fig.4). Training and validation box losses—quantifying localization accuracy through CIoU/DIoU optimization—showed consistent decline across epochs, indicating stable spatial prediction refinement. Classification loss (cross-entropy between predicted genus labels and ground truth) dropped sharply, confirming improved discrimination between coral genera, while parallel training-validation loss trajectories ruled out overfitting. Distribution Focal Loss stabilization near optimal values highlighted reliable confidence score calibration, critical for minimizing false positives in ecological analysis. The model achieved precision=0.75 and recall=0.69 at convergence, balancing false-positive/negative rates essential for accurate abundance estimation of both dominant and rare genera. Performance metrics included mAP@0.5=0.78 (lenient IoU) and mAP@0.5-0.95=0.56, demonstrating reliable detection across varying relevant to ecological monitoring.

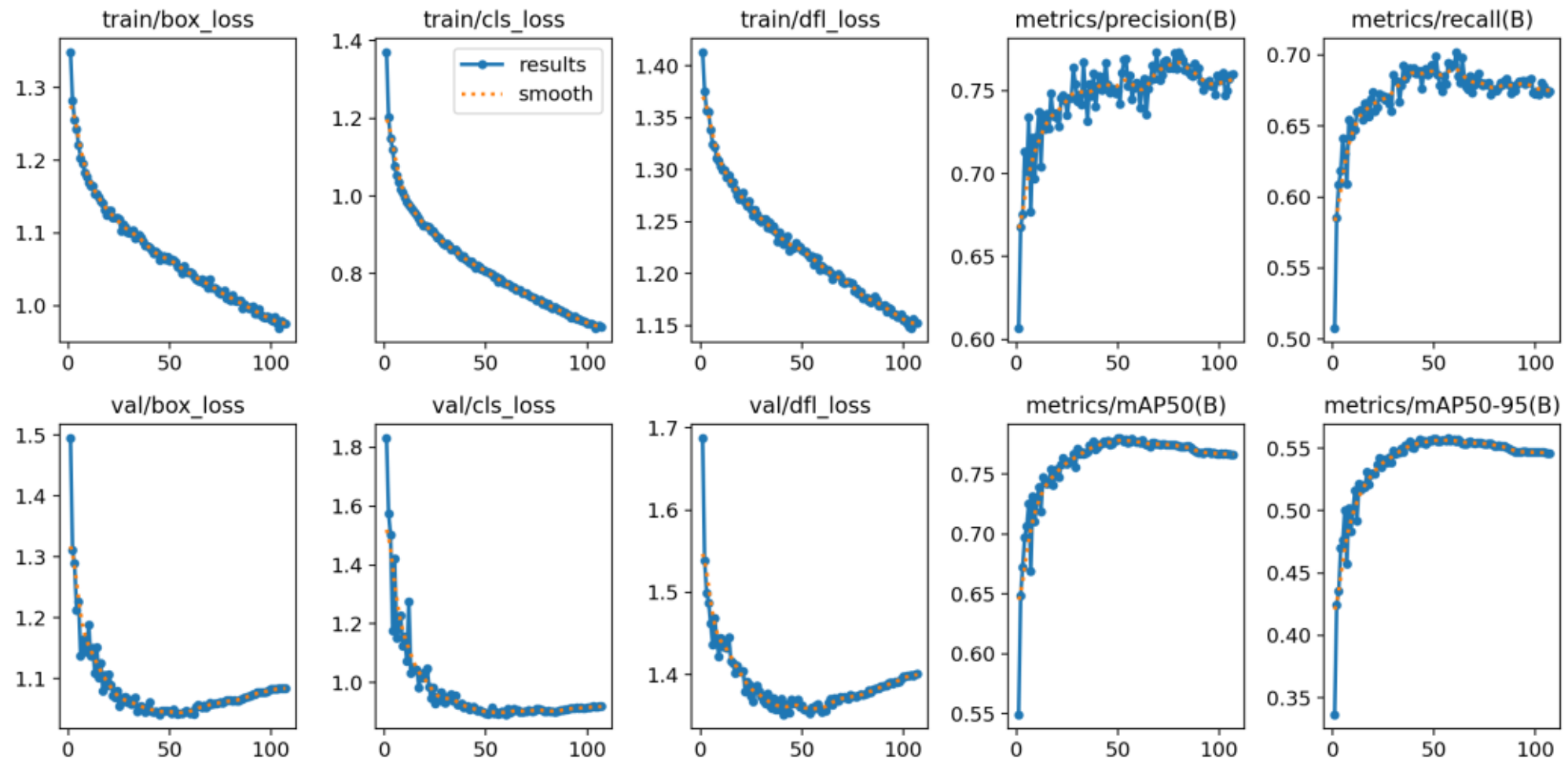

**Figure 4. Object detection model training results**

### 4.2 Segmented result

The trained object detection model was first employed for coral quadrat object detection. By leveraging prior constraints from object detection model's localization outputs, we generated box prompts for the Segment Model, enabling semantic segmentation of coral instances. The segmentation results are illustrated in Figure 5.

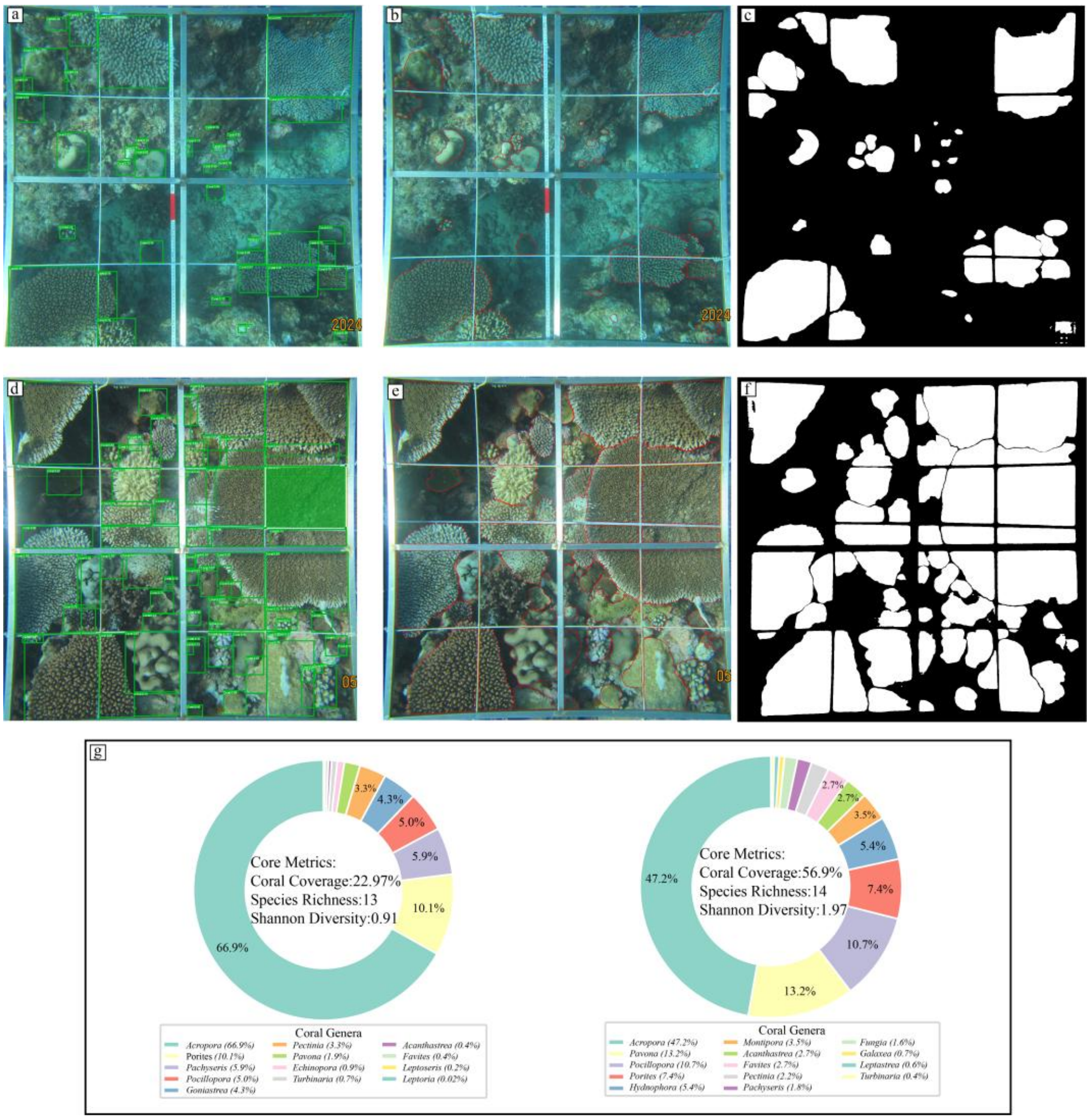

**Figure 5. Semantic Segmentation and Data Extraction Maps**

a: Sparse coral colony object detection visualization; b: Sparse coral segmentation visualization; c: Sparse coral segmentation mask; d: Dense coral aggregation object detection visualization; e: Dense coral aggregation segmentation visualization; f: Dense coral aggregation segmentation mask. g: Data extraction results

### 4.3 Classification and Recognition Module Results

#### 4.3.1 Training results

The rapid decline of the loss curve in the initial phase indicates significant optimization efficacy, followed by stabilization at a low value, which demonstrates consistent model convergence during training. The training accuracy ascends sharply in the early stage and then plateaus near 1.0, reflecting the model's robust learning capacity and the saturation of training performance. The evaluation loss also decreases notably at first, with minor fluctuations (e.g., around 12,500 steps), but ultimately stabilizes at a low level, suggesting strong generalization ability without overfitting. The evaluation accuracy exhibits a steady upward trend, reaching a high plateau (approximately 0.97), consistent with the training accuracy. The consistency between training and evaluation metrics indicates that the model effectively mitigates overfitting and exhibits strong generalization capability on unseen data, thus demonstrating its robustness.

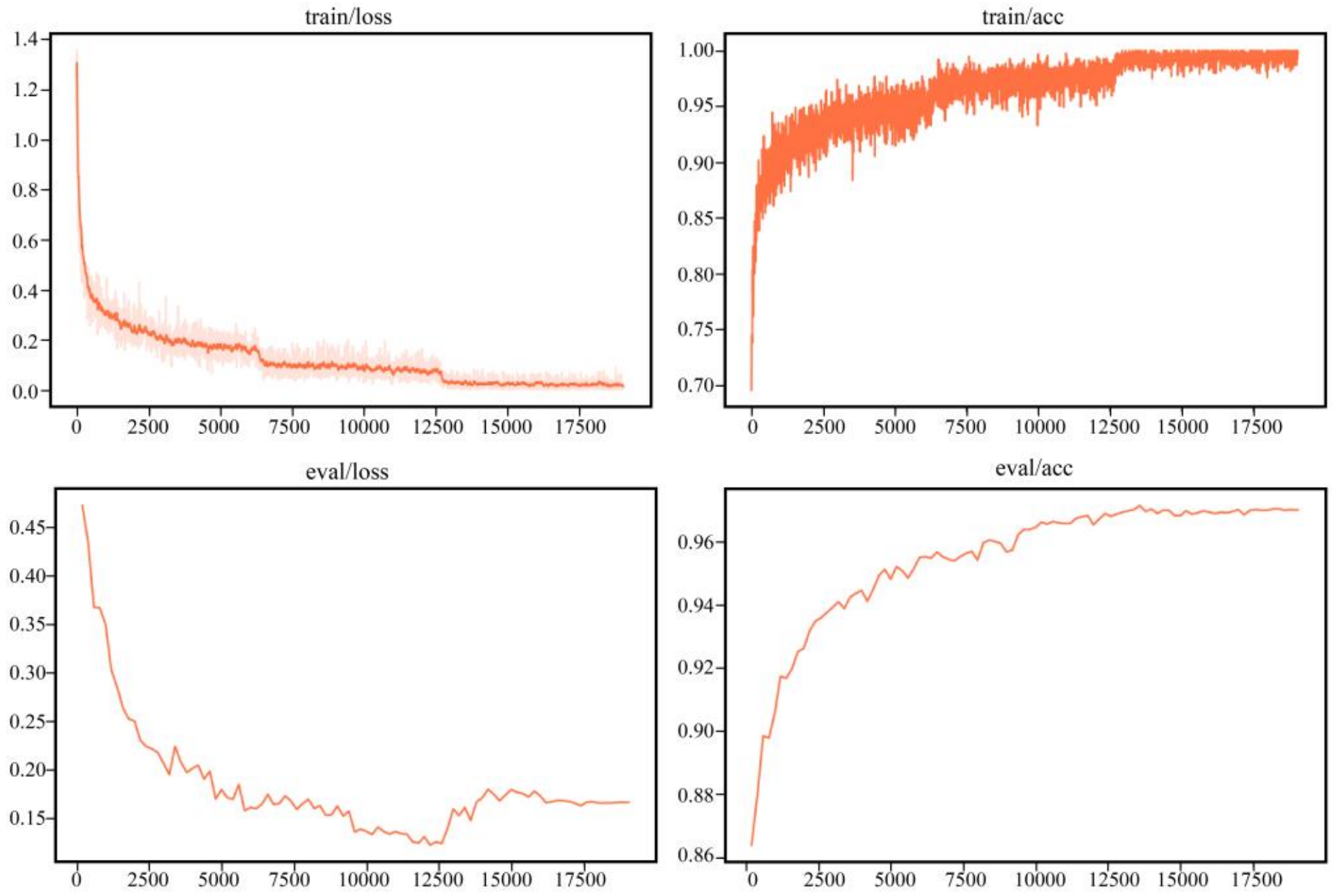

**Figure 6. Multimodal Model Training: Loss and Accuracy Curves**

### 4.3.2 Experiments

Experimental validation confirms a hierarchical improvement in model performance through systematic optimization. Using the baseline model trained on the Core Dataset (9 genera, 9:1 split) as a reference, we first increased LoRA parameters to rank=64 and alpha=128. This adjustment yielded a notable 1.36% accuracy improvement from 79.9% to 81.26%, striking a balance between model expressiveness and computational efficiency. Subsequent full-parameter training further elevated performance to 86.31%, with additional gains achieved through higher input resolution (1020×1024 pixels) and optimized training epochs. When applied to the extended dataset (43 genera + Hybrid), the final configuration attained 88.00% overall accuracy and 85.65% Macro F1-score (Table A.2). Performance stratification revealed 17 genera (e.g., *Acropora*, *Pocillopora*) achieving >90% accuracy, 19 genera (including Hybrid) at 80–90%, and four morphologically convergent genera (*Favites*, *Echinophyllia*) below 70% accuracy. The multimodal framework's superiority stems from its extensive pre-training data and preserved microstructural cues obtained through priori prompts and data fine-tuning. It outperforms traditional methods in accuracy while maintaining balanced precision-recall trade-offs across imbalanced classes.

**Table 2. Experiments Results**

| Objective | Modification | Accuracy (%) | Macro F1 (%) |
|---|---|---|---|
| Baseline model | Top 9 genera, 2%, 9:1 data split | 79.9 | 64.21 |
| Reduce training data | Data split ratio 8:2 | 77.5 | 60.35 |
| Increase data variety | Increased data variety (Top 24 genera, 1%) | 68.87 | 56.66 |
| | Increased data variety (Top 17 genera, | 71.75 | 56.12 |
| | 1.5%) | | |
| Adjust training parameters | Increased one epoch (4) | 80.15 | 65.52 |
| | Decreased one epoch (2) | 78.67 | 60.78 |
| | LoRA parameters rank=64, alpha=128 | 81.26 | 65.30 |
| | Increased image resolution to 1020×1024 | 80.27 | 62.50 |
| | Full Parameter Training | **86.31** | **74.74** |
| | Full Parameter Training with VIT Freezing | 82.0 | 66.11 |
| | Optimal Parameter Combination | **88.00** | **85.65** |

### 4.3.3 Confusion Matrix Analysis

To further analyze the model's performance, we generated a confusion matrix for the optimal model. The overall confusion matrix is shown in Figure 7. This matrix provides a detailed view of the model's classification performance across all coral genera.

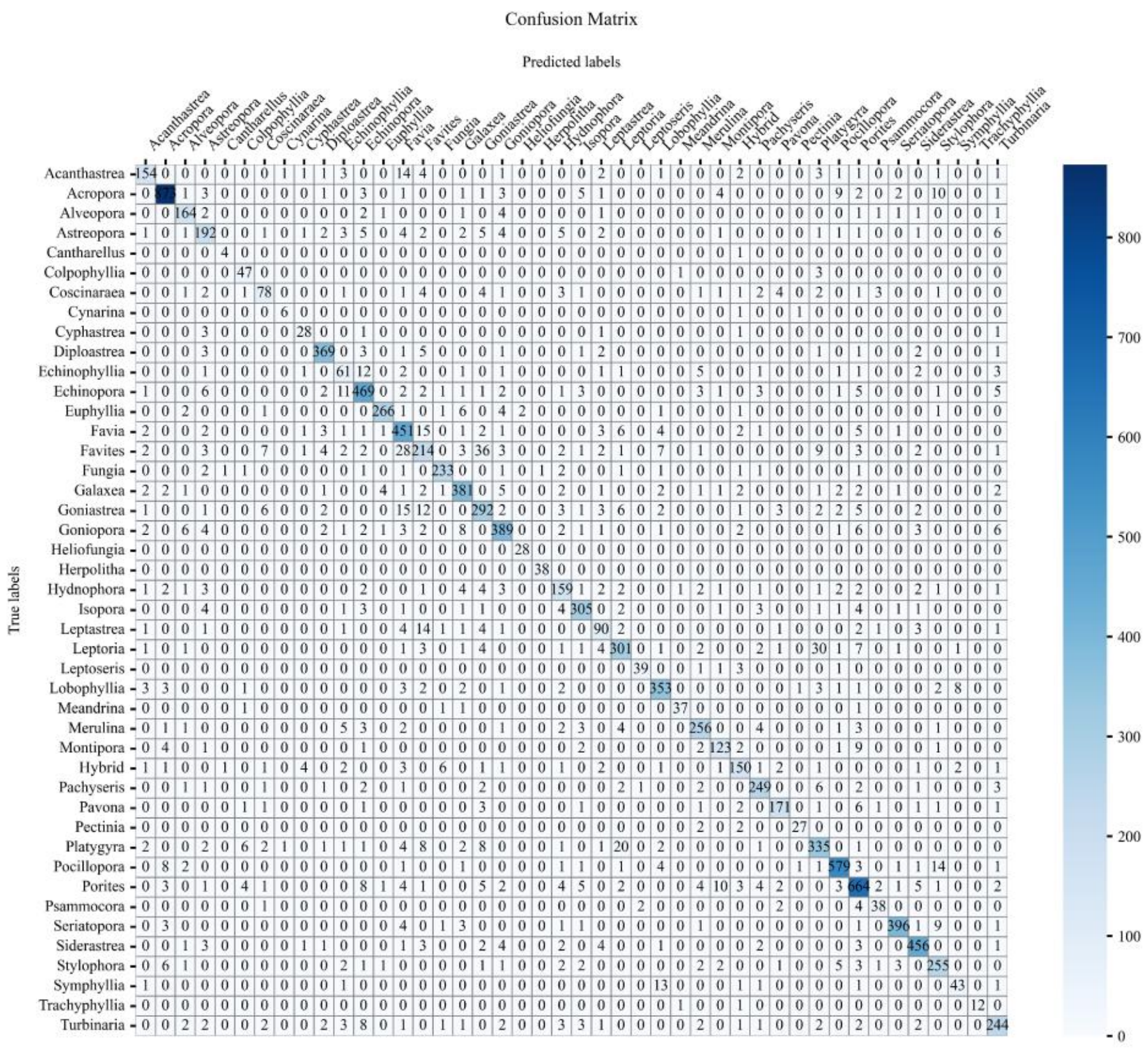

**Figure 7. Confusion matrix for the Multimodal coral identification model**

Additionally, we focused on the confusion matrix for coral genera with lower accuracy rates. This partial confusion matrix is shown in Figure 8. It highlights specific areas where the model struggled, particularly with distinguishing between closely related genera.

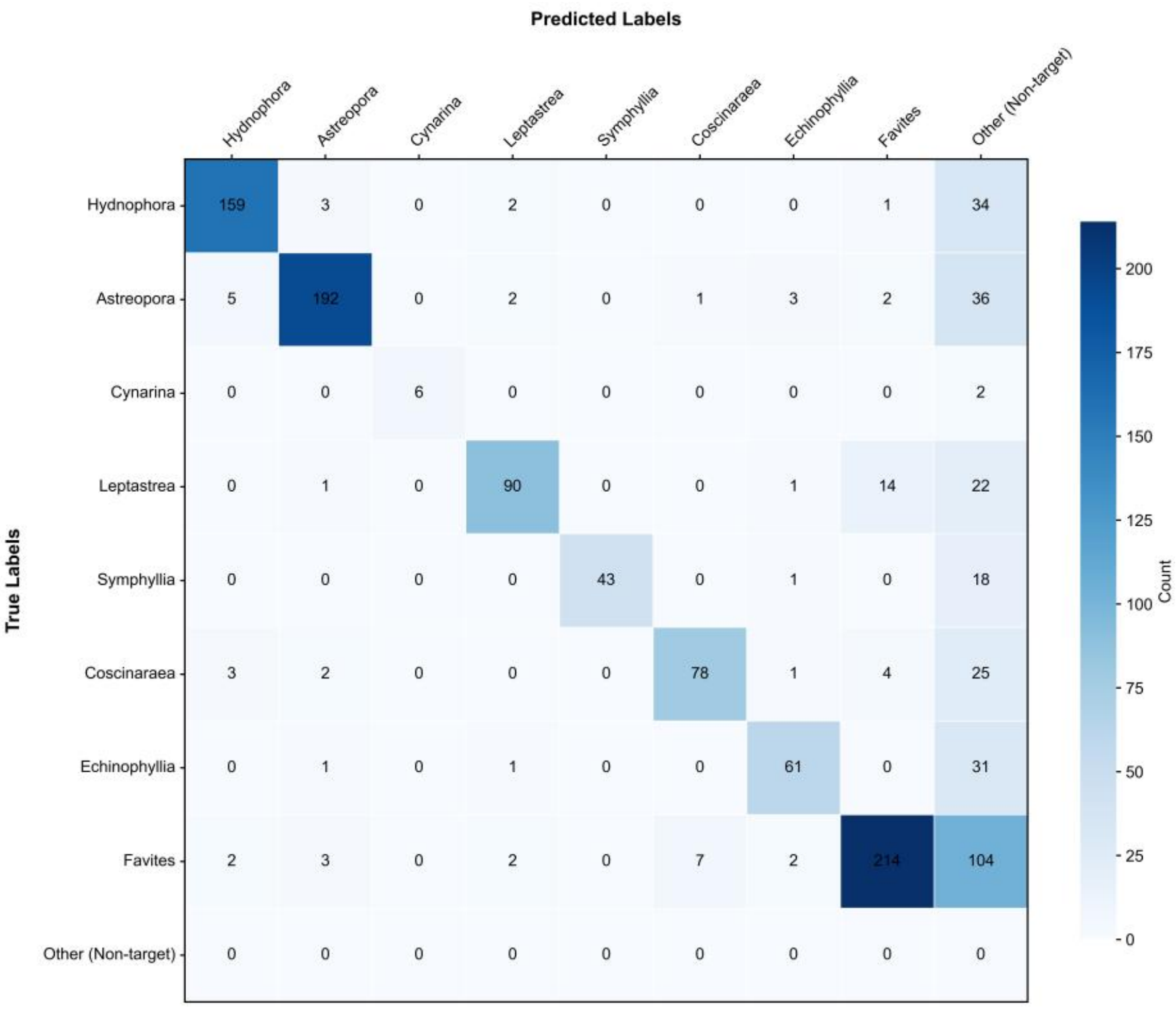

**Figure 8. Partial confusion matrix for coral genera with lower accuracy rates**

The genus with the lowest accuracy was *Favites* (64.07%), with 334 test images. Of these, 28 (8.38%) were misclassified as *Favia*, and 36 (10.78%) as *Goniastrea*. For the *Echinophyllia* genus, there were 94 images in the test set, with 16 images, about 17.02%, being misclassified as *Euphyllia*. In the case of *Symphyllia*, out of 62 images, 16 were misclassified as *Lobophyllia*, accounting for about 25.81%. As for *Coscinaraea*, with 118 images in the test set, there was no clear confusion with other coral genera.

### 4.4 The YH-MINER Data Extraction

The segmented coral images were classified at the genus level using a multimodal coral classification and recognition model. Live coral cover was calculated via segmentation masks, and key ecological indicators were extracted (Fig.5g).

From the coral genus classification results, coral richness was derived. Combined with the coverage of each genus, relative coral abundance was computed. Leveraging both coral coverage and classification outcomes, species diversity indices (Shannon Index, Simpson Index) were further calculated. Dominant coral genera were determined based on their respective coverage values.

## 5. Discussion

### 5.1 Morphological Convergence Leading to Classification Ambiguity

Experimental results show that the recognition accuracy for the genus *Favites* is the lowest

(64.07%), with 8.38% misclassified as *Favia* and 10.78% as *Goniastrea*. This pattern likely stems from their phylogenetic proximity and morphological convergence. Molecular phylogenetic studies place these genera within Clade XVII (family Faviidae), sharing similar corallite wall structures and coenosteal textures (Huang et al., 2011). Both genera exhibit cerioid corallite arrangements with fused walls, while microstructural analyses reveal critical distinctions: *Goniastrea* possesses parathecal walls with abortive septa, contrasting with Favites' synapticulothecal walls lacking these structures. This subtle differentiation in wall architecture, detectable only through skeletal thin-sectioning, likely eludes traditional models. Furthermore, Huang et al. (2014) demonstrated that these genera share subcorallite scale traits including comparable septal dentation patterns and costal cluster densities, which traditional taxonomy overemphasized. Their molecular divergence (4.2-6.7% in ITS sequences) paradoxically exceeds morphological disparity, exemplifying the "coral morphological paradox" where genetic differentiation outpaces observable phenotype changes. These findings align with Budd and Stolarski (2011), who established that Faviidae systematics require integrated micromorphometric analyses to resolve such cryptic complexes.

The confusion between *Echinophyllia* and *Euphyllia* (17.02%) may reflect their shared foliose growth forms and tentacle visibility in living specimens, despite distinct skeletal microstructures (Budd and Stolarski 2011). The high misclassification rate between *Symphyllia* and *Lobophyllia* (25.81%) aligns with their shared brain-coral morphology. Phylogenetic analyses confirm their sister-group relationship within Lobophylliidae, characterized by comparable valley elevation patterns and septal arrangements (Huang et al., 2014). Conversely, *Coscinaraea* showed no significant confusion, likely due to its unique perforated coenosteum and distinct evolutionary lineage within Siderastreidae (Fukami et al., 2008), which provide stronger diagnostic signals. These limitations mirror traditional taxonomic challenges in separating morphologically convergent taxa (Montgomery et al., 2019).

**5.2 Impact of Phylogenetic Proximity on Classification**

The low accuracy for genera like *Favites* (64.07%) and *Echinophyllia* (63.21%) underscores the need for targeted data collection and improved data quality. These results reinforce the necessity of targeted data collection for rare species, as advocated by Andréfouët et al. (2002) in their work on coral connectivity. Rivas et al. (2023) reported significant degradation of coral reefs in the Colombian Caribbean, underscoring the need for models with cross-domain robustness. While our framework improves generalization through multi-source datasets (e.g., CoalNet and GBIF), low accuracy for certain genera highlights the limitations of prior studies that relied on imbalanced data.

Curtis et al. (2024) highlighted the variability and bias in cover estimates derived from seabed images, particularly in sparse, low-cover organisms like cold-water corals. They found that grid-based annotation methods tend to overestimate coral cover, while manual segmentation methods, though more accurate, suffer from size selectivity bias. This aligns with our findings, where manual segmentation of images is recommended to minimize annotator variability and bias. The study by Curtis et al. (2024) also demonstrated that data-driven modeling techniques can reduce uncertainty in cover estimates by addressing size selectivity bias, which is crucial for improving the accuracy of coral monitoring.

### 5.3 System Integration Benefits

Within the coral reef quadrat intelligent analysis system, the cascaded integration of a target detection model and image segmentation model with the multimodal coral identification model enables comprehensive coral reef monitoring, combining precise instance localization via the detection model's robust object detection (achieving mAP=0.78 across varying IoU thresholds) with pixel-level semantic segmentation of coral regions. This framework enhances coral coverage estimation accuracy and facilitates extraction of ecological metrics like density and distribution patterns, generating high-quality segmentation masks even for dense aggregations (Fig.5) to address traditional methods' limitations in quantifying coverage and distinguishing packed colonies. The integration of the MLLM into the coral reef quadrat intelligent analysis system introduces distinct advantages for genus-level classification and ecological feature extraction, complementing the object detection and segmentation modules.

### 5.4 Future Directions

The integration of MLLMs also lays the foundation for multimodal agent-based applications. We have collaborated to attempt integrating large models with underwater robots to support dynamic coral reef health assessment. Such efforts can significantly enhance on-site coral reef analysis capabilities and survey efficiency, enabling timely conservation interventions.

While the YH-MINER have achieved significant success in coral quadrat identification and data extraction tasks, the complexity of coral reef ecosystems necessitates broader research horizons. Coral reefs are complex networks composed of multiple biological communities. To comprehensively assess the health of coral reef ecosystems, future research could further incorporate data from other biological communities within coral quadrats, particularly algal communities and substrates. Additionally, the integration of the coral reef MLLM framework with underwater robotics holds promise for achieving real-time and efficient monitoring of coral reef ecosystems.

**Appendix A:**

**Table A.1: Distribution of coral genera in the extended dataset**

| Genus | Photo Count | Percentage of Total |
|---|---|---|
| *Acropora* | 9215 | 8.08% |
| *Porites* | 7429 | 6.51% |
| *Pocillopora* | 6203 | 5.44% |
| *Echinopora* | 5216 | 4.57% |
| *Favia* | 5032 | 4.41% |
| *Siderastrea* | 4850 | 4.25% |
| *Goniopora* | 4438 | 3.89% |
| *Seriatopora* | 4229 | 3.71% |
| *Galaxea* | 4174 | 3.66% |
| *Platygyra* | 3999 | 3.51% |
| *Diploastrea* | 3902 | 3.42% |
| *Lobophyllia* | 3865 | 3.39% |
| *Leptoria* | 3648 | 3.20% |
| *Goniastrea* | 3619 | 3.17% |
| *Favites* | 3345 | 2.93% |
| *Isopora* | 3343 | 2.93% |
| *Stylophora* | 2899 | 2.54% |
| *Merulina* | 2871 | 2.52% |
| *Euphyllia* | 2863 | 2.51% |
| *Turbinaria* | 2850 | 2.50% |
| *Pachyseris* | 2753 | 2.41% |
| *Fungia* | 2482 | 2.18% |
| *Astreopora* | 2415 | 2.12% |
| *Hydnophora* | 1991 | 1.75% |
| *Pavona* | 1927 | 1.69% |
| *Acanthastrea* | 1916 | 1.68% |
| Hybrid | 1843 | 1.62% |
| *Alveopora* | 1802 | 1.58% |
| *Montipora* | 1460 | 1.28% |
| *Leptastrea* | 1289 | 1.13% |
| *Coscinaraea* | 1130 | 0.99% |
| *Echinophyllia* | 945 | 0.83% |
| *Symphyllia* | 624 | 0.55% |
| *Colpophyllia* | 510 | 0.45% |
| *Psammocora* | 472 | 0.41% |
| *Leptoseris* | 454 | 0.40% |
| *Meandrina* | 413 | 0.36% |
| *Herpolitha* | 386 | 0.34% |
| *Cyphastrea* | 355 | 0.31% |
| *Pectinia* | 318 | 0.28% |
| *Heliofungia* | 284 | 0.25% |
| *Trachyphyllia* | 147 | 0.13% |
| *Cynarina* | 80 | 0.07% |
| *Cantharellus* | 56 | 0.05% |

**Table A.2: The accuracy of Coral identification**

| Genus | Accuracy (%) | Precision (%) | Recall (%) | F1 Score (%) |
|---|---|---|---|---|
| **Overall** | **88.00** | 85.76 | 85.71 | **85.65** |
| *Heliofungia* | 100.00 | 93.33 | 100.00 | 96.55 |
| *Herpolitha* | 100.00 | 97.44 | 100.00 | 98.70 |
| *Acropora* | 94.79 | 96.36 | 94.79 | 95.57 |
| *Diploastrea* | 94.62 | 94.13 | 94.62 | 94.37 |
| *Siderastrea* | 94.02 | 93.63 | 94.02 | 93.83 |
| *Fungia* | 93.95 | 94.72 | 93.95 | 94.33 |
| *Seriatopora* | 93.84 | 97.06 | 93.84 | 95.42 |
| *Pocillopora* | 93.39 | 94.45 | 93.39 | 93.92 |
| *Euphyllia* | 93.01 | 96.73 | 93.01 | 94.83 |
| *Colpophyllia* | 92.16 | 75.81 | 92.16 | 83.19 |
| *Lobophyllia* | 91.45 | 89.59 | 91.45 | 90.51 |
| *Galaxea* | 91.37 | 90.28 | 91.37 | 90.82 |
| *Isopora* | 91.32 | 90.24 | 91.32 | 90.77 |
| *Alveopora* | 91.11 | 88.17 | 91.11 | 89.62 |
| *Pachyseris* | 90.55 | 89.89 | 90.55 | 90.22 |
| *Meandrina* | 90.24 | 92.50 | 90.24 | 91.36 |
| *Echinopora* | 90.02 | 88.32 | 90.02 | 89.16 |
| *Favia* | 89.66 | 81.26 | 89.66 | 85.26 |
| *Porites* | 89.49 | 88.06 | 89.49 | 88.77 |
| *Merulina* | 89.20 | 88.89 | 89.20 | 89.04 |
| *Pavona* | 89.06 | 91.44 | 89.06 | 90.24 |
| *Stylophora* | 88.24 | 85.28 | 88.24 | 86.73 |
| *Goniopora* | 87.81 | 88.81 | 87.81 | 88.31 |
| *Pectinia* | 87.10 | 90.00 | 87.10 | 88.52 |
| *Leptoseris* | 86.67 | 92.86 | 86.67 | 89.66 |
| *Trachyphyllia* | 85.71 | 100.00 | 85.71 | 92.31 |
| *Turbinaria* | 85.61 | 85.61 | 85.61 | 85.61 |
| *Montipora* | 84.25 | 83.67 | 84.25 | 83.96 |
| *Platygyra* | 83.96 | 83.13 | 83.96 | 83.54 |
| *Leptoria* | 82.69 | 85.75 | 82.69 | 84.20 |
| Hybrid | 81.52 | 83.33 | 81.52 | 82.42 |
| *Goniastrea* | 80.89 | 77.66 | 80.89 | 79.24 |
| *Psammocora* | 80.85 | 80.85 | 80.85 | 80.85 |
| *Acanthastrea* | 80.63 | 88.00 | 80.63 | 84.15 |
| *Cyphastrea* | 80.00 | 73.68 | 80.00 | 76.71 |
| *Cantharellus* | 80.00 | 66.67 | 80.00 | 72.73 |
| *Hydnophora* | 79.90 | 78.33 | 79.90 | 79.10 |
| *Astreopora* | 79.67 | 79.34 | 79.67 | 79.50 |
| *Cynarina* | 75.00 | 75.00 | 75.00 | 75.00 |
| *Leptastrea* | 70.31 | 72.58 | 70.31 | 71.43 |
| *Symphyllia* | 69.35 | 79.63 | 69.35 | 74.14 |
| *Coscinaraea* | 69.03 | 76.47 | 69.03 | 72.56 |
| *Echinophyllia* | 64.89 | 61.62 | 64.89 | 63.21 |
| *Favites* | 64.07 | 72.79 | 64.07 | 68.15 |

**Table A.3: Comparison of accuracies obtained by ResNet-50, ResNet-152, DenseNet-121, DenseNet-161**

| Model | Accuracy (%) | Batch Size |
|---|---|---|
| ResNet-50 | 63.90 | 128 |
| ResNet-152 | 68.93 | 128 |
| DenseNet-121 | 69.37 | 128 |
| DenseNet-161 | 70.60 | 64 |